\documentclass[journal=jceaax,manuscript=article]{achemso}
\pdfoutput=1
\usepackage{chemformula}
\usepackage{bm}
\usepackage{natbib}
\usepackage[T1]{fontenc}
\usepackage{subcaption}
\usepackage{amssymb}
\usepackage{multirow}
\usepackage{longtable}
\usepackage{rotating}
\usepackage{hhline}
\usepackage{amsmath}
\usepackage{graphicx}
\usepackage{color}
\usepackage{xcolor}
\usepackage{tikz}
\usepackage{blindtext}
\usepackage{siunitx}
\DeclareSIUnit[quantity-product = ]\percent{\char`\%}
\usetikzlibrary{shapes.geometric}

\author{Dominik~Gond}
\affiliation{Laboratory of Engineering Thermodynamics, RPTU Kaiserslautern, 67663 Kaiserslautern, Germany}

\author{Jan-Tobias~Sohns}
\affiliation{Visual Information Analysis Group, RPTU Kaiserslautern, 67663 Kaiserslautern, Germany}

\author{Heike~Leitte}
\affiliation{Visual Information Analysis Group, RPTU Kaiserslautern, 67663 Kaiserslautern, Germany}

\author{Hans~Hasse}
\affiliation{Laboratory of Engineering Thermodynamics, RPTU Kaiserslautern, 67663 Kaiserslautern, Germany}

\author{Fabian~Jirasek\textsuperscript}
\affiliation{Laboratory of Engineering Thermodynamics, RPTU Kaiserslautern, 67663 Kaiserslautern, Germany}
\email{fabian.jirasek@rptu.de}

\title{Hierarchical Matrix Completion for the Prediction of Properties of Binary Mixtures}

\begin{document}

\begin{abstract}
Predicting the thermodynamic properties of mixtures is crucial for process design and optimization in chemical engineering. Machine learning (ML) methods are gaining increasing attention in this field, but experimental data for training are often scarce, which hampers their application. In this work, we introduce a novel generic approach for improving data-driven models: inspired by the ancient rule "similia similibus solvuntur", we lump components that behave similarly into chemical classes and model them jointly in the first step of a hierarchical approach. While the information on class affiliations can stem in principle from any source, we demonstrate how classes can reproducibly be defined based on mixture data alone by agglomerative clustering. The information from this clustering step is then used as an informed prior for fitting the individual data. We demonstrate the benefits of this approach by applying it in connection with a matrix completion method (MCM) for predicting isothermal activity coefficients at infinite dilution in binary mixtures. Using clustering leads to significantly improved predictions compared to an MCM without clustering. Furthermore, the chemical classes learned from the clustering give exciting insights into what matters on the molecular level for modeling given mixture properties.
\end{abstract}

\section{Introduction}
\label{sec:intro}
Knowledge of the thermodynamic properties of mixtures is crucial for designing and optimizing many industrial processes. However, measuring thermodynamic properties is time-consuming and costly, so that only a tiny fraction of all relevant mixtures have been studied in experiments. Consequently, methods for predicting mixture properties are of paramount importance, and their development is a critical challenge in thermodynamics.

While physical prediction methods for thermodynamic properties have been established for decades \cite{poling2001properties, NRTL, UNIQUAC, UNIFAC1, UNIFAC2, COSMO-RS}, data-driven methods from machine learning (ML) \cite{jirasek2021perspective, medina2022graph, winter2022smile} as well as hybrid approaches combining ML with physical information \cite{jirasek2023combining, specht2024hanna} offer promising alternatives today. Matrix completion methods (MCMs) constitute an interesting class of ML methods for predicting the properties of binary mixtures. These well-established methods in recommender systems have gained much attention through the Netflix Prize \cite{bennett2007netflix} and have recently been transferred to thermodynamics \cite{jirasek2020machine}. The central idea behind using MCMs in thermodynamics is that data on a given thermodynamic property measured for different binary systems can conveniently be stored in a matrix where the rows and columns denote the components of the mixtures, and the entries represent the data of the studied mixture property. Since these properties are usually only available for a small fraction of the mixtures, these matrices are only sparsely occupied with experimental data, and the prediction of unavailable data constitutes a matrix completion problem. MCMs have successfully been developed for predicting activity coefficients\cite{jirasek2020hybridizing, jirasek2022making, damay2021predicting}, Henry's law constants \cite{hayer2022prediction}, and diffusion coefficients \cite{grossmann2022database}. Also, extensions to tensor completion have been proposed \cite{damay2023predicting}.

A drawback of MCMs, and data-driven methods in general, is that their performance depends crucially on the availability of sufficient amounts of  suitable training data. In applications to binary mixtures, they perform poorly for mixtures with components for which only very few training data points are available \cite{jirasek2022making}. One way to address this challenge is to hybridize the MCM with a suitable physical prediction method \cite{jirasek2020hybridizing, jirasek2023prediction, hayerUNIFAC2}, which is, however, only applicable if suitable prediction methods for the property of interest exist.

In this work, we introduce a novel generic approach for improving data-driven methods for predicting mixture properties solely based on the available data without the need for any additional information. The approach is hierarchical: In the first step, similar components are assigned to classes and then jointly modeled using class-specific parameters. In the second step, the learned class-specific parameters are used as informed prior to learning the respective individual component's parameters in a Bayesian framework. This way, prior information on the similarity of components is incorporated into the model so that within a given class, systems that contain components that rarely occur in other systems can profit from systems for which this is not the case. The idea behind this goes back to the ancient rule (established since the days of alchemy): similia similibus solvuntur (Latin), which means that substances that behave similarly are suitable solvents for each other, or -- in our words: can be lumped together. 

The information on component similarities and their affiliation to different classes can stem from any source. The component class affiliations could, e.g., be defined based on readily available pure-component descriptors like dipole moments, based on molecular graph similarities \cite{morgan, similarity}, or also be based on chemical intuition. In this work, we demonstrate that the hierarchical approach even works without any external information. Specifically, we show that component classes can be defined based on mixture data alone using hierarchical agglomerative clustering \cite{mullner2011modern}.

For demonstrating the applicability of our approach, we use the MCM for predicting activity coefficients at infinite dilution $\gamma_{ij}^\infty$ in binary mixtures at \SI[separate-uncertainty = true]{298.15\pm 1}{\kelvin} from our prior work \cite{jirasek2020machine}. However, we emphasize that the approach can be transferred easily to other models.

\section{Database}
\label{sec:data_base}
Experimental data for activity coefficients $\gamma_{ij}^\infty$ of solutes $i$ at infinite dilution in solvents $j$ at \SI[separate-uncertainty = true]{298.15\pm 1}{\kelvin} were taken from the Dortmund Data Bank (DDB) \cite{DortmundDataBank2024}. We used the natural logarithm of the activity coefficient $\ln\gamma_{ij}^\infty$ for computational reasons. Data points labeled as poor quality in the DDB were excluded. Additionally, only solutes and solvents with data for at least two different binary systems were included, as this is necessary for the used leave-one-out analysis detailed below. Moreover, ionic liquids, pure metals, and salts were excluded from the dataset. The arithmetic mean of $\ln\gamma_{ij}^\infty$ was calculated and used for binary mixtures with multiple data points. The final dataset comprises 4\,242 data points and covers $I=238$ solutes and $J=250$ solvents, which can be represented in a sparsely (\SI{7.1}{\percent}) occupied matrix spanned by the solutes and solvents.

\section{Methods}
\label{sec:methods}
\subsection{Overview}

\begin{figure}[H]
    \centering
    \includegraphics[width=\textwidth]{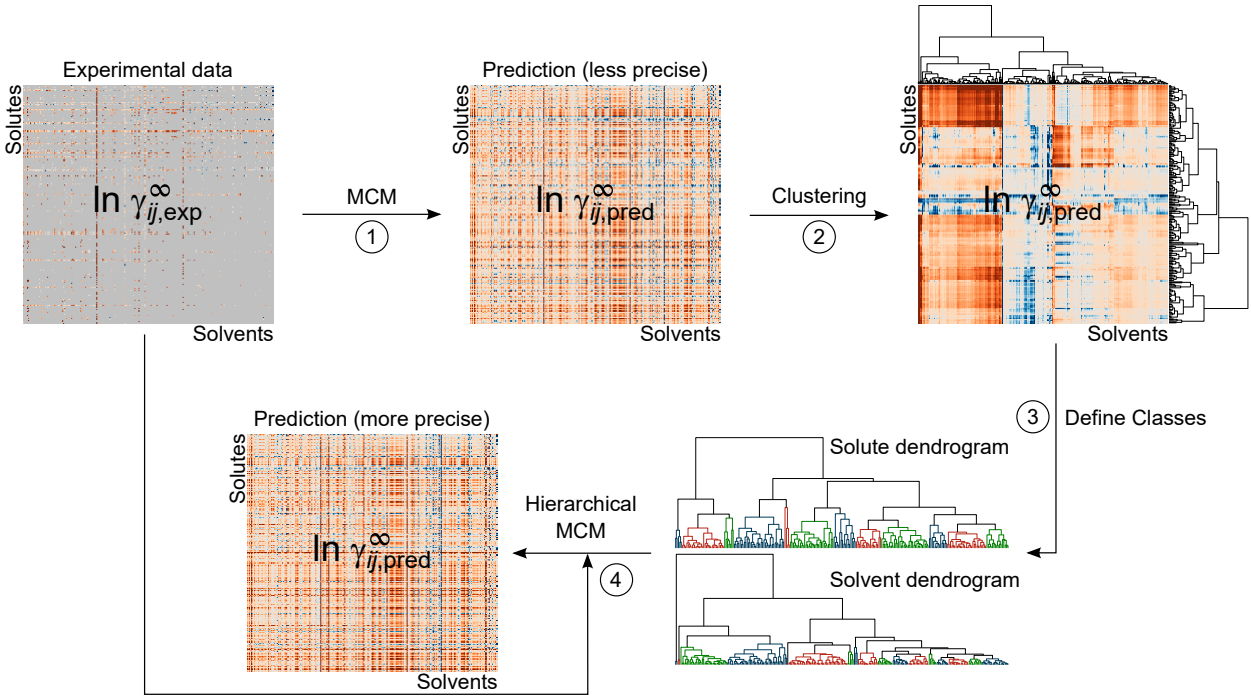}
    \caption{Scheme that illustrates how an MCM can be improved by clustering -- without any additional data. (1) A data-driven MCM is trained on the available experimental data on $\ln \gamma_{ij,\mathrm{exp}}^\infty$. (2) The obtained completed matrix of $\ln \gamma_{ij,\mathrm{pred}}^\infty$ is fed into the agglomerative clustering algorithm. (3) The resulting dendrograms are used to define component classes based on similarity regarding mixture behavior. (4) The class affiliations and experimental data are used in a hierarchical MCM, providing more precise predictions for $\ln \gamma_{ij}^\infty$.}
    \label{fig:scheme_pipeline}
\end{figure}

Figure \ref{fig:scheme_pipeline} gives an overview of our approach. In the first step, a data-driven MCM~\cite{jirasek2020machine} is trained on the available experimental data for $\ln\gamma_{ij}^\infty$, yielding a fully occupied, predicted matrix. This matrix is then fed into a hierarchical agglomerative clustering algorithm. The resulting clustering can be represented in so-called dendrograms for the solutes and the solvents, respectively. These dendrograms encode similarities among the components. Note that a fully occupied matrix is essential for this procedure; otherwise, solutes with missing entries in a given solvent (and vice versa) would be misinterpreted as having similar mixture properties. In the third step, the dendrograms are used to define classes of components and assign each component to its respective class. Finally, the class affiliations of the components and the experimental data are used as input for a hierarchical MCM. We show below that the additional information on the clustering of the components can be used to substantially improve the quality of the predictions compared to using the experimental data alone. 
The following sections provide more details on the individual steps. The approach was evaluated using a leave-one-out analysis, in which the data points considered to be the test data points were consequently omitted from all steps during the training. 

Before we explain the individual steps, we briefly describe the Bayesian MCM used here, which treats parameters as random variables following probability distributions. The MCM does not consider explicit information about the pure components, such as physical descriptors, so it is a collective filtering approach~\cite{ramlatchan2018survey, raghuwanshi2019collaborative}. The MCM solely relies on the available data for binary mixtures from which it infers latent variables (LVs) for all components during the training. In the MCM, $\ln\gamma_{ij}^\infty$ is modeled by the vector product of the two vectors:

\begin{equation}
\label{eq:MCM}
    \ln \gamma_{ij}^\infty = \bm{u}_i^\mathrm{T}\cdot \bm{v}_j\mathrm{,}
\end{equation}

where $\bm{u}_i$ and $\bm{v}_j$ are column vectors of length $K$ containing the learned LVs of the solute $i$ and the solvent $j$, respectively. $K$ is a hyperparameter of the model, which was, however, not varied but set to $K=4$ as in our previous work~\cite{jirasek2020machine}. Normal distributions were chosen as the prior distributions for all parameters of the probabilistic model. The likelihood was defined as a Cauchy distribution, and the MCM was implemented using the probabilistic programming language Stan~\cite{stan}. For inference, we used Gaussian mean-field variational inference. The source code of this MCM can be found in Ref~\cite{jirasek2020machine}.

\subsection{Step 1: Matrix Completion Method}
\label{sub:MCMs}

The first step consists of the direct application of the MCM described above on the experimental data set. As prior normal distributions centered around zero were chosen:

\begin{equation}
    \label{eq:prior_MCM_u}
    p\left( \bm{u}_{ik} \right) = p\left( \bm{v}_{jk} \right) = \mathcal{N} \left(  0,\sigma  \right)\mathrm{\;,\:for\:}k = 1\dots K\mathrm{,}
\end{equation}

where $\sigma$ is the standard deviation, which was fixed at 0.8 as in our prior work~\cite{jirasek2020machine}.
The likelihood was defined as a Cauchy distribution centered around the vector product of the LVs with scale parameter $\lambda = 0.15$ as in our prior work~\cite{jirasek2020machine}:

\begin{equation}
    \label{eq:likelihood_MCM}
    p \left( \ln \gamma_{ij}^\infty \right |\bm{u}_i, \bm{v}_j) = \mathrm{cauchy}\left( \bm{u}_i^\mathrm{T}\cdot \bm{v}_j,\lambda\right){.}
\end{equation}

For brevity, we call this standard MCM (sMCM) here. This step is necessary to fill the matrix so that it can be used for the hierarchical clustering in the next step. 

\subsection{Step 2: Hierarchical Clustering}
\label{sub:hierClus}

The matrix from the sMCM predictions was used as input for hierarchical agglomerative clustering (HAC). For this clustering, each solute $i$ is represented by a vector of length $J=250$ containing $\ln \gamma_{ij}^\infty$ of solute $i$ in all considered solvents predicted by sMCM. Analogously, each solvent $j$ is represented by a vector of length $I=238$ containing $\ln \gamma_{ij}^\infty$ of all considered solutes in the solvent $j$ predicted by sMCM. In other words, each solute $i$ (solvent $j$) is represented by the respective row (column) of the completed matrix  $\ln \gamma_{ij,\mathrm{pred}}^\infty$, cf. Figure \ref{fig:scheme_pipeline}.

In the initial step of the clustering, each component is considered as a distinct cluster. Then, in an iterative procedure, the most similar solutes (solvents) in terms of their above-described vector representation are merged to form a new cluster. This procedure of merging clusters is repeated until all solutes (solvents) are aggregated into a single cluster. The similarity between two solutes (solvents) was defined here as the distance between the two vector representations of the solutes (solvents). In this work, the Euclidean distance was used, and the so-called `complete' linkage criterion~\cite{voorhees1986implementing} was used to define the reference point of each cluster when calculating cluster distances. All definitions related to defining the similarity are hyperparameters of the method. The clustering was performed with the function scipy.cluster.hierarchy.linkage from the python module scipy~\cite{scipy}.

\subsection{Step 3: Defining Classes from Dendrograms}
The result of the hierarchical clustering can be stored in tree-like dendrograms, where similar solutes (solvents) are connected by short paths and dissimilar solutes (solvents) by long paths. Hence, the distance between the initialized clusters (usually represented at the bottom part of the dendrogram) and the position (height) at which two clusters are combined in the dendrogram can be considered as a measure of dissimilarity between clusters of the respective solutes (solvents).
Figure \ref{fig:scheme_pipeline} contains examples of dendrograms for the solutes and the solvents. Based on the obtained dendrograms, classes of solutes (solvents) can be identified in a reproducible, data-driven way. To define these classes, the dendrograms can either be cut at a specific position (dissimilarity), or a certain number of classes can be specified. Either way, the resulting sub-branches can be considered solute (solvent) clusters defined only based on their predicted mixture properties. In this work, we have used the scipy.cluster.hierarchy.fcluster function from the scipy module~\cite{scipy} to obtain a predefined number of solute (solvent) classes and the scipy.cluster.hierarchy.dendrogram function from the same module to visualize the dendrograms. 

\subsection{Step 4: Hierarchical MCM}
\label{sub:hMCM}

In contrast to sMCM, the hierarchical MCM (hMCM) incorporates prior knowledge about the LVs by ruling that the LVs of solutes (solvents) of the same class should be similar. We encode this in the architecture of hMCM by introducing LVs for solute and solvent classes, jointly referred to as component classes in the following, besides the component-specific (solute and solvent-specific) LVs of sMCM.

\begin{figure}[H]
    \centering
    \includegraphics[width=0.9\textwidth]{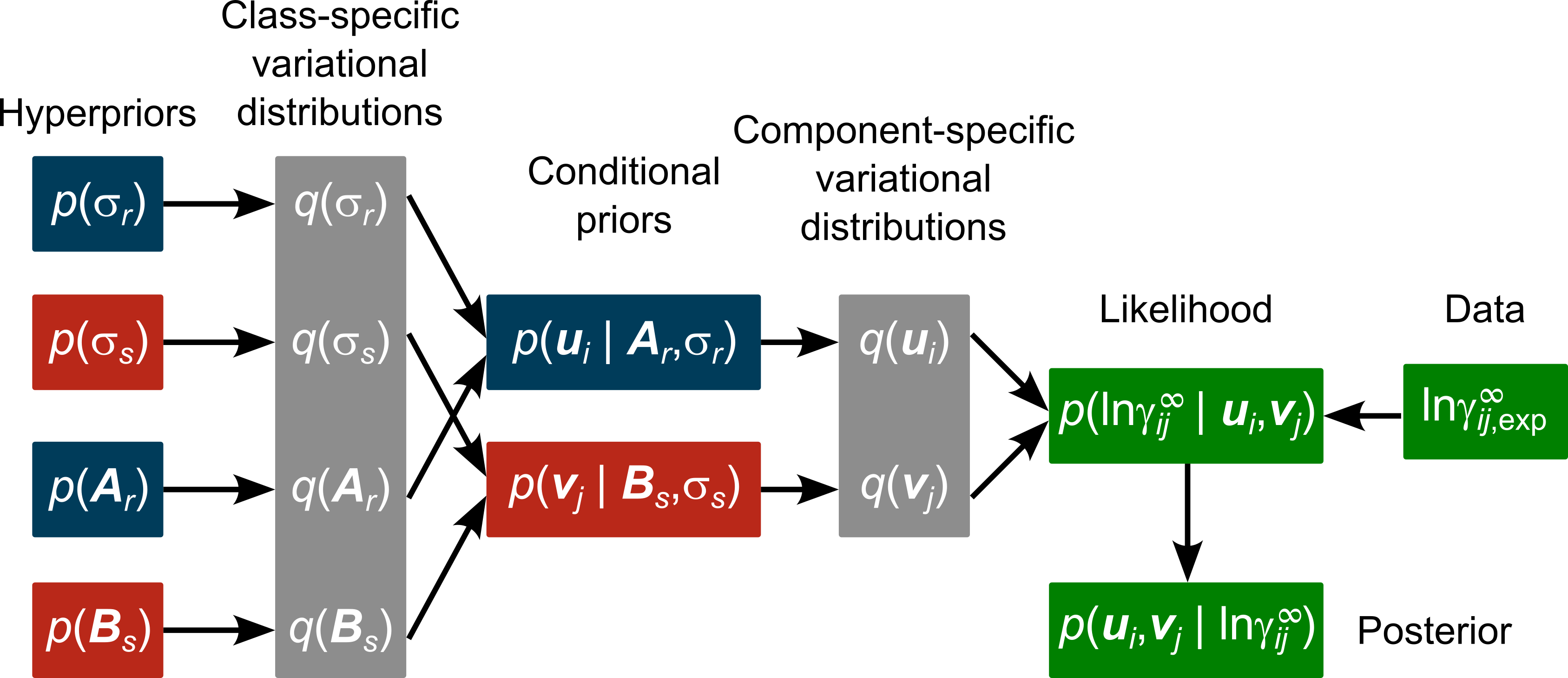}
    \caption{Schematic representation of the hierarchical matrix completion method (hMCM). Class-specific LVs are modeled as variational distributions (grey) drawn from hyperprior distributions for solute (blue) and solvent (red) classes. The class-specific LVs are used to define conditional, class-specific priors, from which component-specific LVs are drawn, analogous to the training of sMCM. The likelihood (green) models how the component-specific LVs explain the training data $\ln\gamma_{ij\mathrm{,exp}}^\infty$ (green). After the training, the posterior distribution (green) for the final component-specific LVs is obtained. $\bm{A}_{r}$ and $\bm{B}_{s}$ denote vectors containing class-specific LVs, $\bm{u}_{i}$ and $\bm{v}_{j}$ denote vectors containing component-specific LVs. The variables are explained in the text.}
    \label{fig:hMCM_scheme}
\end{figure}

Figure \ref{fig:hMCM_scheme} shows a schematic of the hierarchical MCM.
$\bm{A}_r$ and $\bm{B}_s$ denote vectors containing class-specific LVs for the solute class $r$ and the solvent class $s$, respectively. All class-specific LVs are drawn from a hyperprior defined by a normal distribution centered around zero with a fixed standard deviation of $\sigma_\mathrm{hp} = 1$:
\begin{equation}
    \label{eq:prior_hMCM_A}
    p\left( \bm{A}_{rk} \right) =p\left( \bm{B}_{sk} \right) =  \mathcal{N} \left(  0,\sigma_\mathrm{hp}  \right)\mathrm{\;,\:for\:}k = 1\dots K\mathrm{,}
\end{equation}

The component-specific LVs are subsequently sampled from conditional priors modeled as Cauchy distributions of the form:

\begin{equation}
    \label{eq:prior_hMCM_u}
    \begin{aligned}
        p\left( \bm{u}_{ik} | \bm{A}_{rk}, \sigma_r  \right) = \mathrm{Cauchy} \left( \bm{A}_{rk}, \sigma_r \right)\mathrm{\;,\:for\:}k = 1\dots K\mathrm{,} \\
        p\left( \bm{v}_{jk} | \bm{B}_{sk}, \sigma_s  \right) = \mathrm{Cauchy} \left( \bm{B}_{sk}, \sigma_s \right)\mathrm{\;,\:for\:}k = 1\dots K\mathrm{,}
    \end{aligned}
\end{equation}

where the component-specific prior of the solute (solvent) is centered around the respective solute's (solvent's) class-specific LVs. The widths of these prior distributions are defined by class-specific deviation scales $\sigma_r$ and $\sigma_s$ for solute class $r$ and solvent class $s$, respectively, which are drawn from a prior exponential distribution:

\begin{equation}
    \label{eq:prior_stdhMCM_A}
    p\left( \sigma_{r} \right) = p\left( \sigma_{s} \right) = \mathrm{Exponential}\left( \eta \right)\mathrm{,}
\end{equation}

defined by the scale parameter $\eta$, which was set to $\eta=1$ here. The modeling of the component-specific priors as Cauchy distributions was chosen to account for possible outliers in the class assignment, while the flexible inferring of class-specific deviations $\sigma_r$ and $\sigma_s$ accounts for the different homogeneity of the defined classes. With this approach, we ensure that the LVs of a given solute (solvent) are similar to the LVs of other solutes (solvents) from the same class. Therefore, the components implicitly learn from data on all components within its class, sharing statistical strength. The hMCM's likelihood was defined analogously to the sMCM's likelihood (cf. Equation \ref{eq:likelihood_MCM}).

\subsection{Computational Details}
\label{sec:CompDet}
All preprocessing and postprocessing steps were carried out in MATLAB R2021b and Python 3.11, using the modules scipy \cite{scipy} and pandas \cite{pandas}. Both MCMs, sMCM and hMCM, were implemented using the probabilistic programming language Stan~\cite{stan}. The source code for running hMCM in Stan can be found in the Supporting Information (cf. Figure S.1).

The evaluation of the predictive performance was done using a leave-one-out analysis. Hence, the entire pipeline described above, cf. Figure~\ref{fig:scheme_pipeline}, was run 4\,242 times, each time withholding one of the available experimental data points. The withheld data point was then predicted by the respective hMCM, and the prediction was compared to the experimental value. In this way, a fair assessment of the predictive capability of hMCM was ensured.

During model development, $\sigma_\mathrm{hp}$, $\lambda$, and $\eta$ were varied, but we found the model to be highly robust with regard to their variation.

\section{Results and Discussion}
\label{sec:results}

\subsection{Hierarchical Clustering}
\label{sub:res_Clustering}
\begin{figure}[H]
   \centering
   \includegraphics[width=\textwidth]{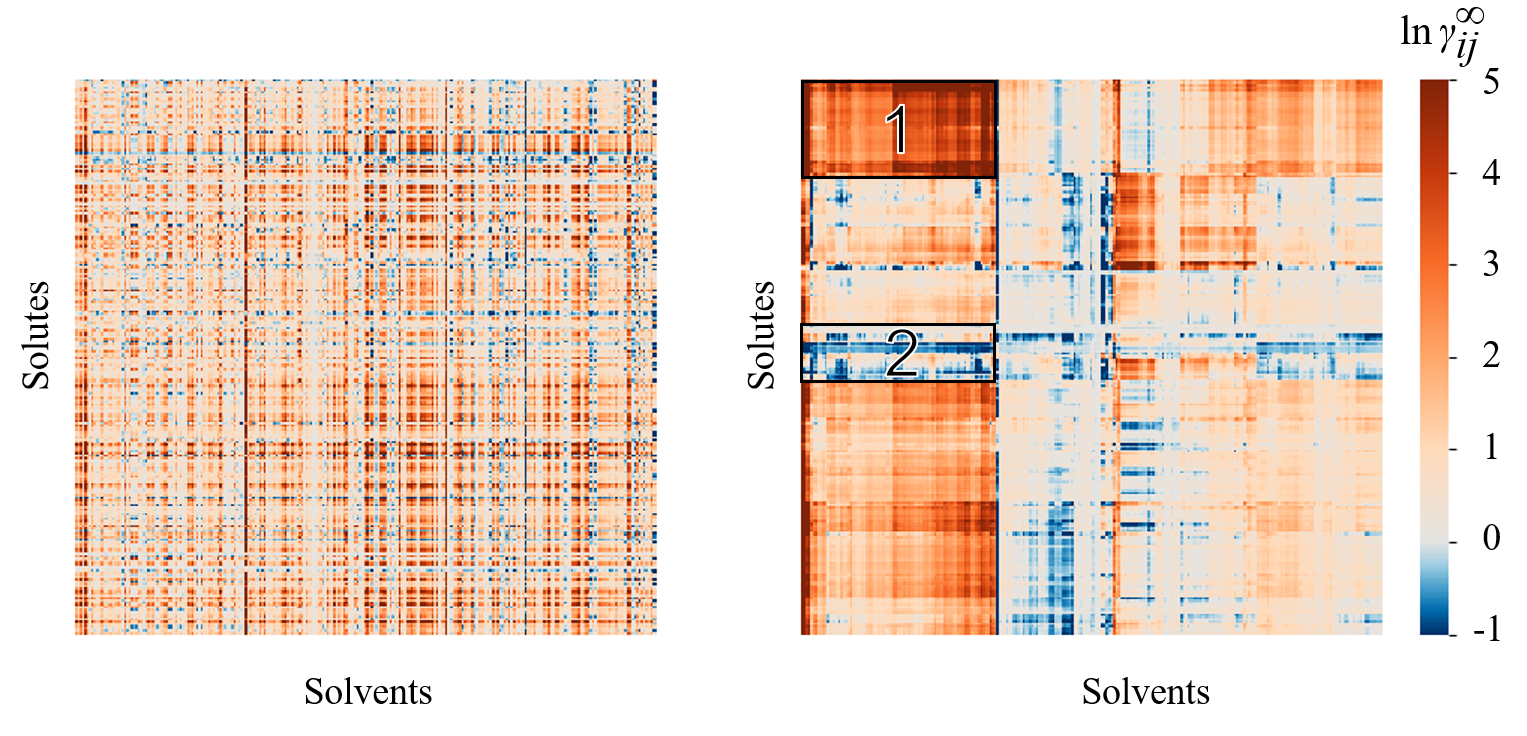}
   \caption{Matrix of $\ln\gamma_{ij}^\infty$ in binary mixtures predicted by sMCM~\cite{jirasek2020machine}. Left: rows (solutes) and columns (solvents) are sorted according to the component identifier from the Dortmund Data Bank (DDB). Right: rows and columns are sorted according to similarities in $\ln\gamma_{ij}^\infty$, obtained by hierarchical agglomerative clustering. The color code indicates the values of $\ln\gamma_{ij}^\infty$. Two distinct blocks in the sorted matrix are marked and discussed in the text as examples.}
   \label{fig:matrices}
\end{figure}

This section discusses the results obtained from the hierarchical clustering of the completed matrix of $\ln\gamma_{ij}^\infty$ predicted with sMCM, which was trained on all available experimental data. Figure \ref{fig:matrices} (left) shows the completed matrix of $\ln\gamma_{ij}^\infty$, with the rows (solutes) and columns (solvents) sorted according to the component identifiers of the DDB. The right part of Figure \ref{fig:matrices} shows the same data after sorting solutes and solvents during the hierarchical clustering. Thereby, components with similar vectors containing $\ln\gamma_{ij}^\infty$, are represented close to each other. In the sorted matrix, block-shaped areas of similar color representing mixtures with similar $\ln\gamma_{ij}^\infty$ can be identified.

Consider, for example, block 1 in Figure~\ref{fig:matrices} (right): it mainly contains dark red entries, indicating very high $\ln\gamma_{ij}^\infty$. This block predominantly comprises mixtures with unfavorable interactions between their components, e.g., aromatic compounds, long-chained esters, and some less polar halogenated hydrocarbons as solutes infinitely diluted in water, polar alcohols, organic acids, or other highly polar components as solvents. The opposite can be seen within block 2 in Figure~\ref{fig:matrices} (right). Here, mixtures of polar alcohols, aldehydes, formamides, and organic acids as solutes with highly polar solvents are found, which exhibit favorable interactions. These examples illustrate that the matrix shown in Figure~\ref{fig:matrices} (right), which is based on the predictions of the data-driven sMCM, is in good agreement with expert knowledge.

\begin{figure}[H]
    \centering
    \includegraphics[width=\textwidth]{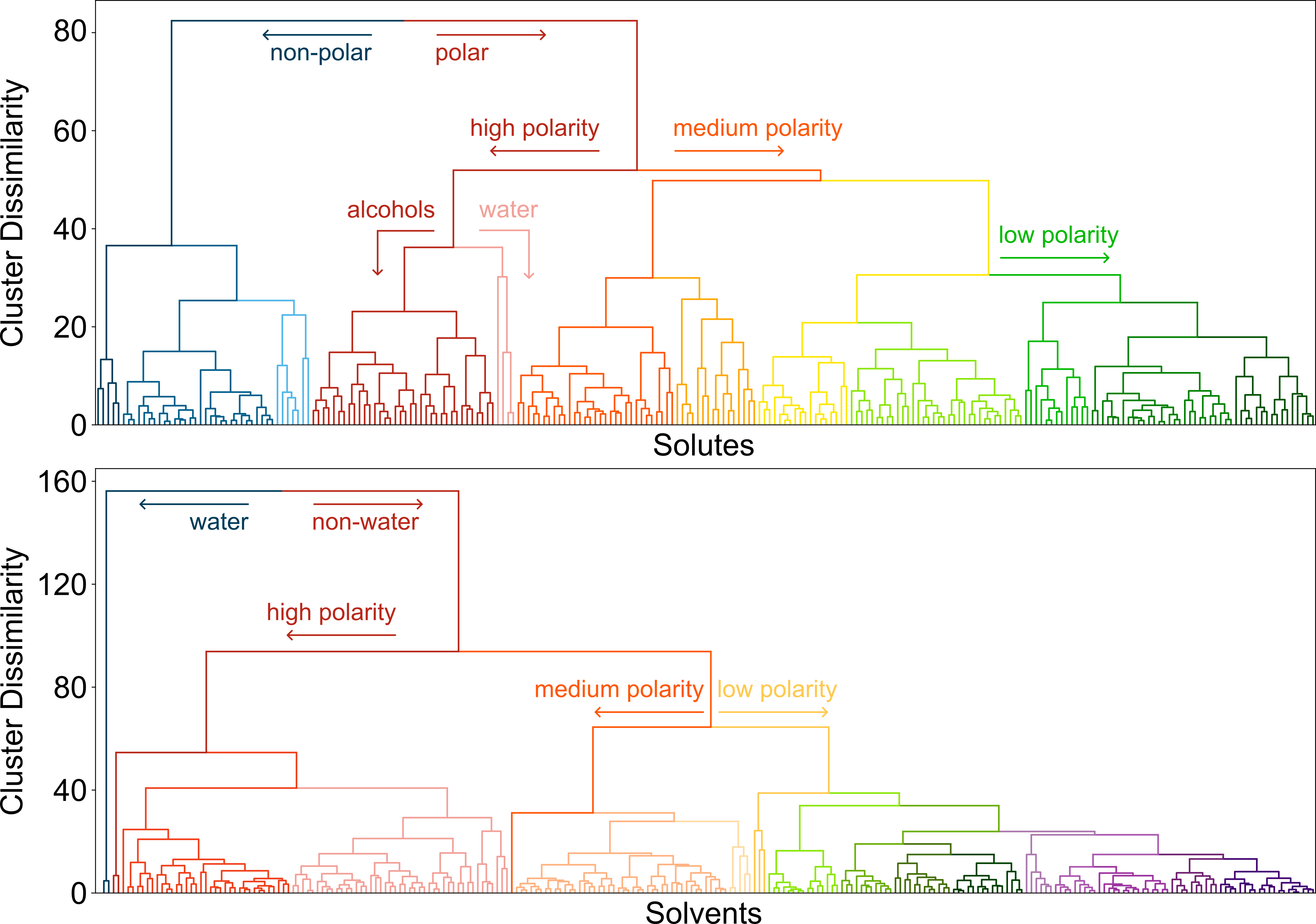}
    \caption{Dendrograms of solutes (top) and solvents (bottom) resulting from the hierarchical clustering based on $\ln\gamma_{ij}^\infty$. The vertical axes represent the dissimilarity between clusters quantified by Euclidean distance and 'complete' linkage, while the horizontal axes show the individual components. Different colors indicate distinctions between clusters, and different color shades indicate the classes defined through visual analysis.}    \label{fig:dendrograms}
\end{figure}

In Figure \ref{fig:dendrograms}, the result of the hierarchical clustering is shown in the form of dendrograms for all studied solutes (top) and all studied solvents (bottom).

For the solutes, the first and most significant distinction (starting from the top of the dendrogram) is the polarity of the components. Very non-polar components, e.g., long-chained alkanes and non-functionalized hydrocarbons (blue), are separated from all other solutes, which consist of components of various polarities. The next distinction made is between components of high polarity (red) and medium to low polarity, where the high polarity branch is further separated into one branch containing short-chained alcohols (dark red) and the other comprised of water, heavy water, and two components with very strong dipole moments (light red). 

For the solvents, the first subdivision is between water and heavy water (blue) and all other components. For the non-water components, the splits according to polarity are similar to what can be seen for the solutes. All this is in good agreement with chemical intuition. The sorted lists of solute and solvent names are shown in the supporting information (cf. Table S.1 and S.2).

\subsection{Hierarchical MCM}
In the following, the results of the developed hMCM for predicting $\ln\gamma_{ij}^\infty$ are shown. These results were obtained using a leave-one-out analysis to ensure a fair assessment based on true predictions (cf. section 'Computational Details'). The number of solute classes was set to 12 and that of solvent classes to 17, based on the visual analysis of the dendrograms in Figure \ref{fig:dendrograms}. 

\begin{figure}[H]
    \centering
    \includegraphics[width=0.8\textwidth]{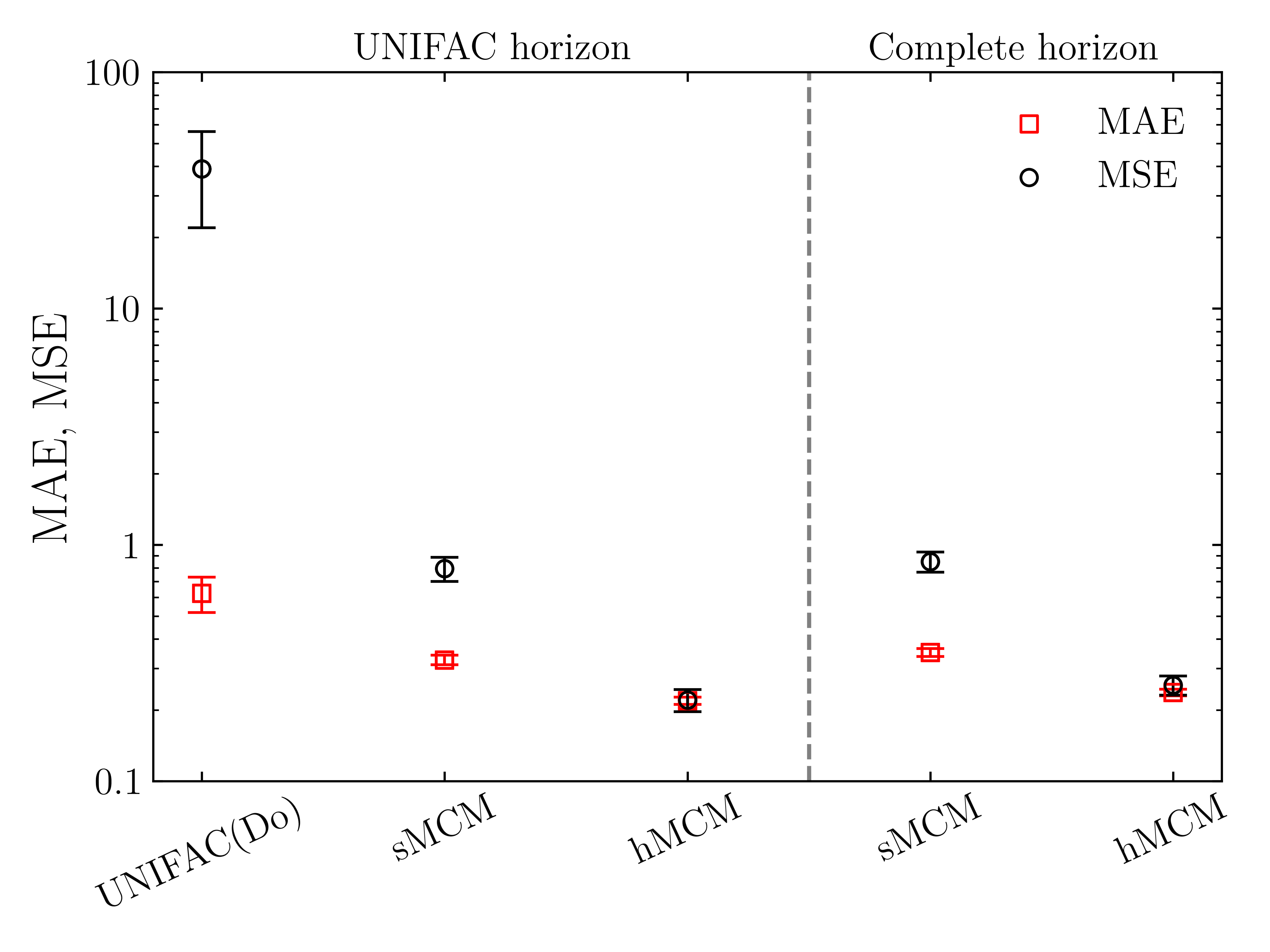}
    \caption{Mean absolute error (MAE) and mean squared error (MSE) of the developed hMCM for predicting $\ln\gamma_{ij}^\infty$ at \SI{298.15}{\kelvin} in binary mixtures and comparison to the data-driven sMCM~\cite{jirasek2020machine} and the physical gold standard UNIFAC(Do)~\cite{modUNIFAC}. The complete horizon covers our complete data set; the UNIFAC horizon includes only the mixtures that UNIFAC(Do) can describe. Error bars indicate the standard errors of the means.}    \label{fig:result_errors}
\end{figure}

Figure \ref{fig:result_errors} summarizes the performance of hMCM in terms of mean average error (MAE) and mean squared error (MSE) of the predictions for the complete data set obtained from the leave-one-out analysis. For comparison, also the scores of the data-driven sMCM~\cite{jirasek2020machine} and the state-of-the-art physical model for predicting activity coefficients, modified UNIFAC (Dortmund)~\cite{modUNIFAC}, abbreviated as UNIFAC(Do) in the following, are shown. Since UNIFAC(Do) can not predict all data from the present data set due to missing parameters, the comparison is made on two different horizons: While the complete horizon covers all 4\,242 binary mixtures for which experimental data are available, the UNIFAC horizon, which UNIFAC(Do) can also describe, comprises 3\,396 of the 4\,242 binary mixtures (\SI{80.0}{\percent} of our data set).

The results in Figure \ref{fig:result_errors} show that hMCM performs significantly better than both sMCM~\cite{jirasek2020machine} and UNIFAC(Do)~\cite{modUNIFAC}. The improvement compared to sMCM is particularly astonishing since both models are based on precisely the same training data. The results clearly show that the hierarchical model structure of hMCM substantially improves the efficiency of using the data during the training. We emphasize that this improvement is not caused by an increased flexibility of hMCM compared to the sMCM since, ultimately, both models use the same number of component-specific parameters (LVs) for making predictions. The additional class-specific parameters introduced in hMCM can be understood as regularizers that "smoothen" the parameter differences within solute and solvent classes. Both sMCM and hMCM clearly outperform UNIFAC(Do), which is remarkable considering that UNIFAC(Do) has assumedly been trained on most of the data points, such that the comparison is biased in favor of UNIFAC(Do). 

\begin{figure}[H]
    \centering
    \includegraphics[width=\textwidth]{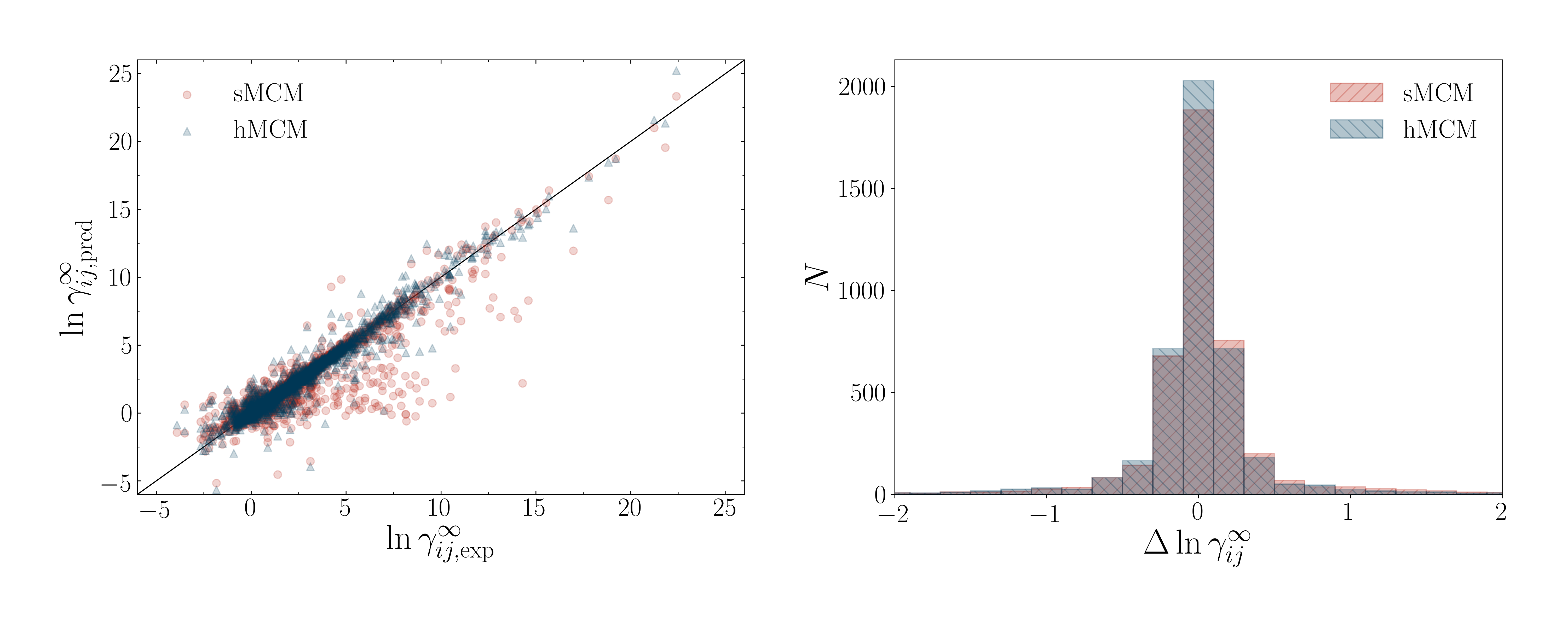}
    \caption{Comparison of the predicted (pred) $\ln\gamma_{ij}^\infty$ in binary mixtures at \SI{298.15}{\kelvin} with sMCM (red) and the developed hMCM (blue) with the respective experimental (exp) test data. Left: parity plot of predictions over experimental data. Right: histogram of the deviations of the predictions from the experimental data, where $N$ is the number of binary mixtures predicted with a deviation $\Delta\ln\gamma_{ij}^\infty = \ln\gamma_{ij\mathrm{,exp}}^\infty - \ln\gamma_{ij\mathrm{,pred}}^\infty$. The displayed interval in the histogram contains \SI{96.4}{\percent} (sMCM) and \SI{98.6}{\percent} (hMCM) of all considered data points.}    \label{fig:result_parity}\end{figure}

In Figure \ref{fig:result_parity}, the predictions with sMCM and hMCM are compared in a parity plot (left) and a histogram of the deviations of the predictions from the experimental data (right).

In both representations, it is evident that hMCM improves the prediction accuracy, especially for those mixtures that sMCM describes rather poorly, i.e., the hierarchical approach significantly reduces the number of severe outliers of sMCM. Most notable is the reduced number of underestimated experimental values by sMCM in Figure \ref{fig:result_parity}(left), which mainly comprises mixtures containing water as the solvent and exotic, poorly measured solutes (measured in less than three solvents), accounting for most of the improvement of hMCM over sMCM. All this clearly underlines the regularizing effect of using information from the component clustering in the hierarchical approach.
\section{Conclusions}
\label{sec:conclusion}

In this work, we present a novel generic approach for improving data-driven models for predicting the properties of mixtures. The approach is hierarchical and based on an old principle already known to alchemists as “similia similibus solvuntur”, which we reinterpret here in the following way: if we have a mixture of two components (A + B), and we have two other components A’ and B’ where A’ is similar to A, and B’ is similar to B, then also the mixture (A’ + B’) will be similar to (A + B). In other words, for the prediction of mixture properties, A and A’ can be lumped into a component class, and so can 
B and B’. Of course, the usefulness of this evident rule depends on the interpretation of the term “similar”. 

We have demonstrated the benefits that can be drawn from this idea using a matrix completion method (MCM) as an example and by applying a certain concept of similarity, but we emphasize that the idea can equally be applied to other data-driven ML methods and in connection with other concepts of similarity.

In the first step of our hierarchical approach, similar components are modeled together as a class of components, and class-specific model parameters are learned from the data. These class-specific parameters are subsequently used as informed prior for learning component-specific model parameters in a Bayesian framework. Our hypothesis is that this enhances the model predictions compared to the corresponding approach without explicitly using the similarity. Improvements are in particular expected regarding the reduction of outliers, as by the lumping of similar components, the predictions for systems with components A for which the database is narrow can benefit from data on systems with similar components A’, for which more data are available. If no additional component descriptors are used for measuring the similarity of A and A’ (i.e., if the concept of similarity is defined based on the existing data alone), the similarity-based method and the corresponding original method have exactly the same database -- and the benefits of the similarity-based approach come for free.

We have demonstrated the feasibility of this approach and its benefits by applying it to an MCM for the prediction of activity coefficients $\ln\gamma_{ij}^\infty$ of solutes $i$ infinitely diluted in solvents $j$  at \SI{298.15}{\kelvin}, which is a problem that we had already tackled without the similarity approach in previous work~\cite{jirasek2020machine}. We have demonstrated that suitable class assignments can be inferred directly from the mixture data using hierarchical agglomerative clustering without any additional information on the components. However, as the matrix with the experimental data on $\ln\gamma_{ij}^\infty$ is only sparsely occupied, it is difficult to use it directly for defining these similarities. In the first step of our hierarchical approach, we have therefore completed the matrix with a standard MCM (sMCM), as in our previous work~\cite{jirasek2020machine}. The similarity of two solutes (solvents) was then defined based on the corresponding vectors that are given by the entries of the rows (columns) of the full matrix, using a simple Euclidean distance, and classes of the solutes and solvents were defined based on dendrograms. The resulting hierarchical MCM (hMCM) significantly improves the accuracy compared to the sMCM and also outperforms the state-of-the-art model, modified UNIFAC (Dortmund). 

In future work, the similarity-based hierarchical approach should be applied to other ML models for predicting mixture properties. Moreover, it can also be used to obtain estimates for mixtures of components not included in the training data by using the learned class-specific parameters as approximate component-specific parameters of unknown components, overcoming a notable limitation of sMCM.

\section*{Acknowledgments}
The authors gratefully acknowledge financial support by Carl Zeiss Foundation in the frame of the project ‘Process Engineering 4.0’ and by DFG in the frame of the Priority Program SPP2363 'Molecular Machine Learning' (grant number 497201843). Furthermore, FJ gratefully acknowledges financial support by DFG in the frame of the Emmy-Noether program (grant number 528649696). 
\clearpage
\bibliography{literature.bib}

\providecommand{\latin}[1]{#1}
\makeatletter
\providecommand{\doi}
  {\begingroup\let\do\@makeother\dospecials
  \catcode`\{=1 \catcode`\}=2 \doi@aux}
\providecommand{\doi@aux}[1]{\endgroup\texttt{#1}}
\makeatother
\providecommand*\mcitethebibliography{\thebibliography}
\csname @ifundefined\endcsname{endmcitethebibliography}
  {\let\endmcitethebibliography\endthebibliography}{}
\begin{mcitethebibliography}{33}
\providecommand*\natexlab[1]{#1}
\providecommand*\mciteSetBstSublistMode[1]{}
\providecommand*\mciteSetBstMaxWidthForm[2]{}
\providecommand*\mciteBstWouldAddEndPuncttrue
  {\def\EndOfBibitem{\unskip.}}
\providecommand*\mciteBstWouldAddEndPunctfalse
  {\let\EndOfBibitem\relax}
\providecommand*\mciteSetBstMidEndSepPunct[3]{}
\providecommand*\mciteSetBstSublistLabelBeginEnd[3]{}
\providecommand*\EndOfBibitem{}
\mciteSetBstSublistMode{f}
\mciteSetBstMaxWidthForm{subitem}{(\alph{mcitesubitemcount})}
\mciteSetBstSublistLabelBeginEnd
  {\mcitemaxwidthsubitemform\space}
  {\relax}
  {\relax}

\bibitem[Poling \latin{et~al.}(2001)Poling, Prausnitz, O'connell,
  \latin{et~al.} others]{poling2001properties}
Poling,~B.~E.; Prausnitz,~J.~M.; O'connell,~J.~P.; others \emph{The properties
  of gases and liquids}; Mcgraw-hill New York, 2001; Vol.~5\relax
\mciteBstWouldAddEndPuncttrue
\mciteSetBstMidEndSepPunct{\mcitedefaultmidpunct}
{\mcitedefaultendpunct}{\mcitedefaultseppunct}\relax
\EndOfBibitem
\bibitem[Renon and Prausnitz(1968)Renon, and Prausnitz]{NRTL}
Renon,~H.; Prausnitz,~J.~M. Local compositions in thermodynamic excess
  functions for liquid mixtures. \emph{AIChE journal} \textbf{1968}, \emph{14},
  135--144\relax
\mciteBstWouldAddEndPuncttrue
\mciteSetBstMidEndSepPunct{\mcitedefaultmidpunct}
{\mcitedefaultendpunct}{\mcitedefaultseppunct}\relax
\EndOfBibitem
\bibitem[Abrams and Prausnitz(1975)Abrams, and Prausnitz]{UNIQUAC}
Abrams,~D.~S.; Prausnitz,~J.~M. Statistical thermodynamics of liquid mixtures:
  a new expression for the excess Gibbs energy of partly or completely miscible
  systems. \emph{AIChE journal} \textbf{1975}, \emph{21}, 116--128\relax
\mciteBstWouldAddEndPuncttrue
\mciteSetBstMidEndSepPunct{\mcitedefaultmidpunct}
{\mcitedefaultendpunct}{\mcitedefaultseppunct}\relax
\EndOfBibitem
\bibitem[Fredenslund \latin{et~al.}(1975)Fredenslund, Jones, and
  Prausnitz]{UNIFAC1}
Fredenslund,~A.; Jones,~R.~L.; Prausnitz,~J.~M. Group-contribution estimation
  of activity coefficients in nonideal liquid mixtures. \emph{AIChE Journal}
  \textbf{1975}, \emph{21}, 1086--1099\relax
\mciteBstWouldAddEndPuncttrue
\mciteSetBstMidEndSepPunct{\mcitedefaultmidpunct}
{\mcitedefaultendpunct}{\mcitedefaultseppunct}\relax
\EndOfBibitem
\bibitem[Fredenslund(2012)]{UNIFAC2}
Fredenslund,~A. \emph{Vapor-liquid equilibria using UNIFAC: a
  group-contribution method}; Elsevier, 2012\relax
\mciteBstWouldAddEndPuncttrue
\mciteSetBstMidEndSepPunct{\mcitedefaultmidpunct}
{\mcitedefaultendpunct}{\mcitedefaultseppunct}\relax
\EndOfBibitem
\bibitem[Klamt(1995)]{COSMO-RS}
Klamt,~A. Conductor-like screening model for real solvents: a new approach to
  the quantitative calculation of solvation phenomena. \emph{The Journal of
  Physical Chemistry} \textbf{1995}, \emph{99}, 2224--2235\relax
\mciteBstWouldAddEndPuncttrue
\mciteSetBstMidEndSepPunct{\mcitedefaultmidpunct}
{\mcitedefaultendpunct}{\mcitedefaultseppunct}\relax
\EndOfBibitem
\bibitem[Jirasek and Hasse(2021)Jirasek, and Hasse]{jirasek2021perspective}
Jirasek,~F.; Hasse,~H. Perspective: machine learning of thermophysical
  properties. \emph{Fluid Phase Equilibria} \textbf{2021}, \emph{549},
  113206\relax
\mciteBstWouldAddEndPuncttrue
\mciteSetBstMidEndSepPunct{\mcitedefaultmidpunct}
{\mcitedefaultendpunct}{\mcitedefaultseppunct}\relax
\EndOfBibitem
\bibitem[Medina \latin{et~al.}(2022)Medina, Linke, Stoll, and
  Sundmacher]{medina2022graph}
Medina,~E. I.~S.; Linke,~S.; Stoll,~M.; Sundmacher,~K. Graph neural networks
  for the prediction of infinite dilution activity coefficients. \emph{Digital
  Discovery} \textbf{2022}, \emph{1}, 216--225\relax
\mciteBstWouldAddEndPuncttrue
\mciteSetBstMidEndSepPunct{\mcitedefaultmidpunct}
{\mcitedefaultendpunct}{\mcitedefaultseppunct}\relax
\EndOfBibitem
\bibitem[Winter \latin{et~al.}(2022)Winter, Winter, Schilling, and
  Bardow]{winter2022smile}
Winter,~B.; Winter,~C.; Schilling,~J.; Bardow,~A. A smile is all you need:
  predicting limiting activity coefficients from SMILES with natural language
  processing. \emph{Digital Discovery} \textbf{2022}, \emph{1}, 859--869\relax
\mciteBstWouldAddEndPuncttrue
\mciteSetBstMidEndSepPunct{\mcitedefaultmidpunct}
{\mcitedefaultendpunct}{\mcitedefaultseppunct}\relax
\EndOfBibitem
\bibitem[Jirasek and Hasse(2023)Jirasek, and Hasse]{jirasek2023combining}
Jirasek,~F.; Hasse,~H. Combining machine learning with physical knowledge in
  thermodynamic modeling of fluid mixtures. \emph{Annual Review of Chemical and
  Biomolecular Engineering} \textbf{2023}, \emph{14}, 31--51\relax
\mciteBstWouldAddEndPuncttrue
\mciteSetBstMidEndSepPunct{\mcitedefaultmidpunct}
{\mcitedefaultendpunct}{\mcitedefaultseppunct}\relax
\EndOfBibitem
\bibitem[Specht \latin{et~al.}(2024)Specht, Nagda, Fellenz, Mandt, Hasse, and
  Jirasek]{specht2024hanna}
Specht,~T.; Nagda,~M.; Fellenz,~S.; Mandt,~S.; Hasse,~H.; Jirasek,~F. HANNA:
  Hard-constraint Neural Network for Consistent Activity Coefficient
  Prediction. \emph{arXiv preprint arXiv:2407.18011} \textbf{2024}, \relax
\mciteBstWouldAddEndPunctfalse
\mciteSetBstMidEndSepPunct{\mcitedefaultmidpunct}
{}{\mcitedefaultseppunct}\relax
\EndOfBibitem
\bibitem[Bennett \latin{et~al.}(2007)Bennett, Lanning, \latin{et~al.}
  others]{bennett2007netflix}
Bennett,~J.; Lanning,~S.; others The netflix prize. Proceedings of KDD cup and
  workshop. 2007; p~35\relax
\mciteBstWouldAddEndPuncttrue
\mciteSetBstMidEndSepPunct{\mcitedefaultmidpunct}
{\mcitedefaultendpunct}{\mcitedefaultseppunct}\relax
\EndOfBibitem
\bibitem[Jirasek \latin{et~al.}(2020)Jirasek, Alves, Damay, Vandermeulen,
  Bamler, Bortz, Mandt, Kloft, and Hasse]{jirasek2020machine}
Jirasek,~F.; Alves,~R.~A.; Damay,~J.; Vandermeulen,~R.~A.; Bamler,~R.;
  Bortz,~M.; Mandt,~S.; Kloft,~M.; Hasse,~H. Machine learning in
  thermodynamics: Prediction of activity coefficients by matrix completion.
  \emph{The journal of physical chemistry letters} \textbf{2020}, \emph{11},
  981--985\relax
\mciteBstWouldAddEndPuncttrue
\mciteSetBstMidEndSepPunct{\mcitedefaultmidpunct}
{\mcitedefaultendpunct}{\mcitedefaultseppunct}\relax
\EndOfBibitem
\bibitem[Jirasek \latin{et~al.}(2020)Jirasek, Bamler, and
  Mandt]{jirasek2020hybridizing}
Jirasek,~F.; Bamler,~R.; Mandt,~S. Hybridizing physical and data-driven
  prediction methods for physicochemical properties. \emph{Chemical
  Communications} \textbf{2020}, \emph{56}, 12407--12410\relax
\mciteBstWouldAddEndPuncttrue
\mciteSetBstMidEndSepPunct{\mcitedefaultmidpunct}
{\mcitedefaultendpunct}{\mcitedefaultseppunct}\relax
\EndOfBibitem
\bibitem[Jirasek \latin{et~al.}(2022)Jirasek, Bamler, Fellenz, Bortz, Kloft,
  Mandt, and Hasse]{jirasek2022making}
Jirasek,~F.; Bamler,~R.; Fellenz,~S.; Bortz,~M.; Kloft,~M.; Mandt,~S.;
  Hasse,~H. Making thermodynamic models of mixtures predictive by machine
  learning: matrix completion of pair interactions. \emph{Chemical Science}
  \textbf{2022}, \emph{13}, 4854--4862\relax
\mciteBstWouldAddEndPuncttrue
\mciteSetBstMidEndSepPunct{\mcitedefaultmidpunct}
{\mcitedefaultendpunct}{\mcitedefaultseppunct}\relax
\EndOfBibitem
\bibitem[Damay \latin{et~al.}(2021)Damay, Jirasek, Kloft, Bortz, and
  Hasse]{damay2021predicting}
Damay,~J.; Jirasek,~F.; Kloft,~M.; Bortz,~M.; Hasse,~H. Predicting activity
  coefficients at infinite dilution for varying temperatures by matrix
  completion. \emph{Industrial \& Engineering Chemistry Research}
  \textbf{2021}, \emph{60}, 14564--14578\relax
\mciteBstWouldAddEndPuncttrue
\mciteSetBstMidEndSepPunct{\mcitedefaultmidpunct}
{\mcitedefaultendpunct}{\mcitedefaultseppunct}\relax
\EndOfBibitem
\bibitem[Hayer \latin{et~al.}(2022)Hayer, Jirasek, and
  Hasse]{hayer2022prediction}
Hayer,~N.; Jirasek,~F.; Hasse,~H. Prediction of Henry's law constants by matrix
  completion. \emph{AIChE Journal} \textbf{2022}, \emph{68}, e17753\relax
\mciteBstWouldAddEndPuncttrue
\mciteSetBstMidEndSepPunct{\mcitedefaultmidpunct}
{\mcitedefaultendpunct}{\mcitedefaultseppunct}\relax
\EndOfBibitem
\bibitem[Gro{\ss}mann \latin{et~al.}(2022)Gro{\ss}mann, Bellaire, Hayer,
  Jirasek, and Hasse]{grossmann2022database}
Gro{\ss}mann,~O.; Bellaire,~D.; Hayer,~N.; Jirasek,~F.; Hasse,~H. Database for
  liquid phase diffusion coefficients at infinite dilution at 298 K and matrix
  completion methods for their prediction. \emph{Digital Discovery}
  \textbf{2022}, \emph{1}, 886--897\relax
\mciteBstWouldAddEndPuncttrue
\mciteSetBstMidEndSepPunct{\mcitedefaultmidpunct}
{\mcitedefaultendpunct}{\mcitedefaultseppunct}\relax
\EndOfBibitem
\bibitem[Damay \latin{et~al.}(2023)Damay, Ryzhakov, Jirasek, Hasse, Oseledets,
  and Bortz]{damay2023predicting}
Damay,~J.; Ryzhakov,~G.; Jirasek,~F.; Hasse,~H.; Oseledets,~I.; Bortz,~M.
  Predicting Temperature-Dependent Activity Coefficients at Infinite Dilution
  Using Tensor Completion. \emph{Chemie Ingenieur Technik} \textbf{2023},
  \emph{95}, 1061--1069\relax
\mciteBstWouldAddEndPuncttrue
\mciteSetBstMidEndSepPunct{\mcitedefaultmidpunct}
{\mcitedefaultendpunct}{\mcitedefaultseppunct}\relax
\EndOfBibitem
\bibitem[Jirasek \latin{et~al.}(2023)Jirasek, Hayer, Abbas, Schmid, and
  Hasse]{jirasek2023prediction}
Jirasek,~F.; Hayer,~N.; Abbas,~R.; Schmid,~B.; Hasse,~H. Prediction of
  parameters of group contribution models of mixtures by matrix completion.
  \emph{Physical Chemistry Chemical Physics} \textbf{2023}, \emph{25},
  1054--1062\relax
\mciteBstWouldAddEndPuncttrue
\mciteSetBstMidEndSepPunct{\mcitedefaultmidpunct}
{\mcitedefaultendpunct}{\mcitedefaultseppunct}\relax
\EndOfBibitem
\bibitem[Hayer \latin{et~al.}(2024)Hayer, Wendel, Mandt, Hasse, and
  Jirasek]{hayerUNIFAC2}
Hayer,~N.; Wendel,~T.; Mandt,~S.; Hasse,~H.; Jirasek,~F. Advancing
  Thermodynamic Group-Contribution Methods by Machine Learning: UNIFAC 2.0.
  2024; \url{https://arxiv.org/abs/2408.05220}\relax
\mciteBstWouldAddEndPuncttrue
\mciteSetBstMidEndSepPunct{\mcitedefaultmidpunct}
{\mcitedefaultendpunct}{\mcitedefaultseppunct}\relax
\EndOfBibitem
\bibitem[Morgan(1965)]{morgan}
Morgan,~H.~L. The generation of a unique machine description for chemical
  structures-a technique developed at chemical abstracts service. \emph{Journal
  of chemical documentation} \textbf{1965}, \emph{5}, 107--113\relax
\mciteBstWouldAddEndPuncttrue
\mciteSetBstMidEndSepPunct{\mcitedefaultmidpunct}
{\mcitedefaultendpunct}{\mcitedefaultseppunct}\relax
\EndOfBibitem
\bibitem[Willett \latin{et~al.}(1998)Willett, Barnard, and Downs]{similarity}
Willett,~P.; Barnard,~J.~M.; Downs,~G.~M. Chemical similarity searching.
  \emph{Journal of chemical information and computer sciences} \textbf{1998},
  \emph{38}, 983--996\relax
\mciteBstWouldAddEndPuncttrue
\mciteSetBstMidEndSepPunct{\mcitedefaultmidpunct}
{\mcitedefaultendpunct}{\mcitedefaultseppunct}\relax
\EndOfBibitem
\bibitem[M{\"u}llner(2011)]{mullner2011modern}
M{\"u}llner,~D. Modern hierarchical, agglomerative clustering algorithms.
  \emph{arXiv preprint arXiv:1109.2378} \textbf{2011}, \relax
\mciteBstWouldAddEndPunctfalse
\mciteSetBstMidEndSepPunct{\mcitedefaultmidpunct}
{}{\mcitedefaultseppunct}\relax
\EndOfBibitem
\bibitem[{Dortmund Data Bank}()]{DortmundDataBank2024}
{Dortmund Data Bank} {, 2024}. \url{www.ddbst.com}\relax
\mciteBstWouldAddEndPuncttrue
\mciteSetBstMidEndSepPunct{\mcitedefaultmidpunct}
{\mcitedefaultendpunct}{\mcitedefaultseppunct}\relax
\EndOfBibitem
\bibitem[Ramlatchan \latin{et~al.}(2018)Ramlatchan, Yang, Liu, Li, Wang, and
  Li]{ramlatchan2018survey}
Ramlatchan,~A.; Yang,~M.; Liu,~Q.; Li,~M.; Wang,~J.; Li,~Y. A survey of matrix
  completion methods for recommendation systems. \emph{Big Data Mining and
  Analytics} \textbf{2018}, \emph{1}, 308--323\relax
\mciteBstWouldAddEndPuncttrue
\mciteSetBstMidEndSepPunct{\mcitedefaultmidpunct}
{\mcitedefaultendpunct}{\mcitedefaultseppunct}\relax
\EndOfBibitem
\bibitem[Raghuwanshi and Pateriya(2019)Raghuwanshi, and
  Pateriya]{raghuwanshi2019collaborative}
Raghuwanshi,~S.~K.; Pateriya,~R. Collaborative filtering techniques in
  recommendation systems. \emph{Data, Engineering and Applications: Volume 1}
  \textbf{2019}, 11--21\relax
\mciteBstWouldAddEndPuncttrue
\mciteSetBstMidEndSepPunct{\mcitedefaultmidpunct}
{\mcitedefaultendpunct}{\mcitedefaultseppunct}\relax
\EndOfBibitem
\bibitem[Kucukelbir \latin{et~al.}(2015)Kucukelbir, Ranganath, Gelman, and
  Blei]{stan}
Kucukelbir,~A.; Ranganath,~R.; Gelman,~A.; Blei,~D. Automatic variational
  inference in Stan. \emph{Advances in neural information processing systems}
  \textbf{2015}, \emph{28}\relax
\mciteBstWouldAddEndPuncttrue
\mciteSetBstMidEndSepPunct{\mcitedefaultmidpunct}
{\mcitedefaultendpunct}{\mcitedefaultseppunct}\relax
\EndOfBibitem
\bibitem[Voorhees(1986)]{voorhees1986implementing}
Voorhees,~E.~M. Implementing agglomerative hierarchic clustering algorithms for
  use in document retrieval. \emph{Information Processing \& Management}
  \textbf{1986}, \emph{22}, 465--476\relax
\mciteBstWouldAddEndPuncttrue
\mciteSetBstMidEndSepPunct{\mcitedefaultmidpunct}
{\mcitedefaultendpunct}{\mcitedefaultseppunct}\relax
\EndOfBibitem
\bibitem[Virtanen \latin{et~al.}(2020)Virtanen, Gommers, Oliphant, Haberland,
  Reddy, Cournapeau, Burovski, Peterson, Weckesser, Bright, {van der Walt},
  Brett, Wilson, Millman, Mayorov, Nelson, Jones, Kern, Larson, Carey, Polat,
  Feng, Moore, {VanderPlas}, Laxalde, Perktold, Cimrman, Henriksen, Quintero,
  Harris, Archibald, Ribeiro, Pedregosa, {van Mulbregt}, and {SciPy 1.0
  Contributors}]{scipy}
Virtanen,~P. \latin{et~al.}  {{SciPy} 1.0: Fundamental Algorithms for
  Scientific Computing in Python}. \emph{Nature Methods} \textbf{2020},
  \emph{17}, 261--272\relax
\mciteBstWouldAddEndPuncttrue
\mciteSetBstMidEndSepPunct{\mcitedefaultmidpunct}
{\mcitedefaultendpunct}{\mcitedefaultseppunct}\relax
\EndOfBibitem
\bibitem[pandas~development team(2020)]{pandas}
pandas~development team,~T. pandas-dev/pandas: Pandas. 2020;
  \url{https://doi.org/10.5281/zenodo.3509134}\relax
\mciteBstWouldAddEndPuncttrue
\mciteSetBstMidEndSepPunct{\mcitedefaultmidpunct}
{\mcitedefaultendpunct}{\mcitedefaultseppunct}\relax
\EndOfBibitem
\bibitem[Constantinescu and Gmehling(2016)Constantinescu, and
  Gmehling]{modUNIFAC}
Constantinescu,~D.; Gmehling,~J. Further development of modified UNIFAC
  (Dortmund): revision and extension 6. \emph{Journal of Chemical \&
  Engineering Data} \textbf{2016}, \emph{61}, 2738--2748\relax
\mciteBstWouldAddEndPuncttrue
\mciteSetBstMidEndSepPunct{\mcitedefaultmidpunct}
{\mcitedefaultendpunct}{\mcitedefaultseppunct}\relax
\EndOfBibitem
\end{mcitethebibliography}
\end{document}


%
%
%
%
%


\section{Stan Code for the Hierarchical MCM}
\label{sec:intro}

\begin{figure}[H]
    \centering
    \includegraphics[width=\textwidth]{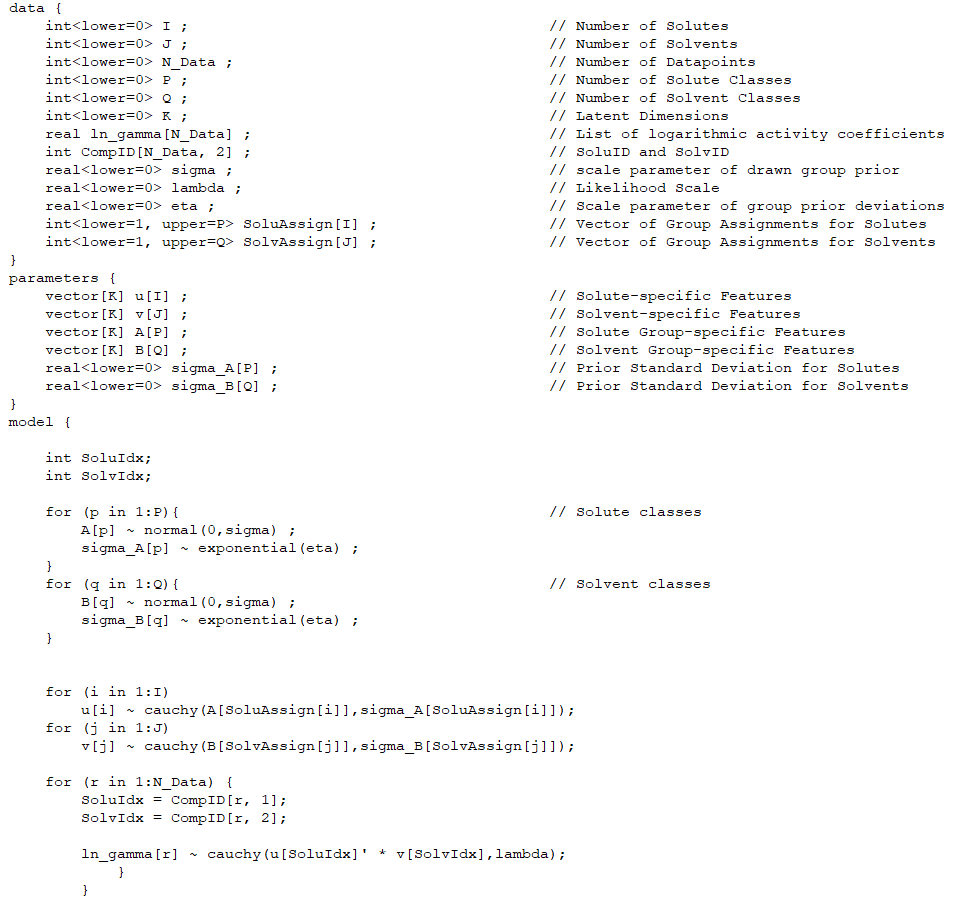}
    \caption{Stan code of the probabilistic generative model for the prediction of limiting activity coefficients by the developed hierarchical MCM.}
    \label{fig:hMCM_CODE}
\end{figure}
\newpage

\section{Class Assignments of Solutes and Solvents}
The following two tables list the components used in this work, ordered according to the respective dendrograms, together with the number of their assigned component class.
\begin{longtable}{ll}
\caption{DDB name and class number of solutes, obtained from hierarchical clustering on the completed matrix of $\ln\gamma_{ij}^\infty$ predicted by sMCM.} \label{tab:long_solu} \\
\hline
\textbf{DDB Name} & \textbf{Class Number} \\
\hline
\endfirsthead

\multicolumn{2}{c}%
{{\bfseries \tablename\ \thetable{} -- continued from previous page}} \\
\hline
\textbf{DDB Name} & \textbf{Class Number} \\
\hline
\endhead

\hline \multicolumn{2}{r}{{Continued on next page}} \\ \hline
\endfoot

\hline \hline
\endlastfoot

Eicosane & 0 \\
Hexadecane & 0 \\
Tetraethylstannane & 0 \\
Dodecane & 0 \\
n-Undecane & 0 \\
Octane & 1 \\
2,2,4-Trimethylpentane & 1 \\
2,5-Dimethylhexane & 1 \\
2,2-Dimethylpentane & 1 \\
2-Methylpentane & 1 \\
Cyclooctane & 1 \\
Tetramethylstannane & 1 \\
2-Methylhexane & 1 \\
Heptane & 1 \\
2,3,4-Trimethyl pentane & 1 \\
Ethylcyclohexane & 1 \\
1-Octene & 1 \\
trans-1,4-Dimethylcyclohexane & 1 \\
2,4-Dimethylpentane & 1 \\
Hexane & 1 \\
Iodobenzene & 1 \\
2-Methylbutane & 1 \\
Cycloheptane & 1 \\
2,2-Dimethylbutane & 1 \\
Pentane & 1 \\
2,3-Dimethylbutane & 1 \\
2,3-Dimethylpentane & 1 \\
2,4,4-Trimethyl-1-pentene & 1 \\
1-Heptene & 1 \\
Methylcyclohexane & 1 \\
3-Methylpentane & 1 \\
Methylcyclopentane & 1 \\
cis-2-Hexene & 1 \\
Cyclohexane & 1 \\
1-Hexene & 1 \\
1-Phenyldodecane & 2 \\
Decane & 2 \\
n-Butylcyclohexane & 2 \\
1-Decene & 2 \\
Nonane & 2 \\
p-Terphenyl & 2 \\
Chrysene & 2 \\
1-Butanol & 3 \\
2-Methyl-1-propanol & 3 \\
2-Butanol & 3 \\
tert-Butanol & 3 \\
N-Methylcaprolactam & 3 \\
2-Propanol & 3 \\
1-Propanol & 3 \\
Acetonitrile & 3 \\
2,2,2-Trifluoroethanol & 3 \\
Nitromethane & 3 \\
Phenol & 3 \\
Perfluoro-n-heptane & 3 \\
Propionitrile & 3 \\
Nitroethane & 3 \\
Butanenitrile & 3 \\
Isobutyronitrile & 3 \\
2,5-Dimethylpyrazine & 3 \\
Cyclohexylamine & 3 \\
tert-Pentanol & 3 \\
Propionic acid & 3 \\
Perfluorohexane & 3 \\
1-Hexanol & 3 \\
1-Pentanol & 3 \\
Cyclohexanol & 3 \\
3-Methyl-1-butanol & 3 \\
Aniline & 3 \\
2-Methylphenol & 3 \\
3-Methylphenol & 3 \\
4-Methylphenol & 3 \\
1-Heptanol & 3 \\
1-Octanol & 3 \\
2-Octanol & 3 \\
1-Nitropropane & 3 \\
Pentanenitrile & 3 \\
Benzonitrile & 3 \\
Hexanenitrile & 3 \\
Water & 4 \\
Deuterium oxide <Heavy water> & 4 \\
N-Methylformamide & 4 \\
Sulfolane & 4 \\
Tetrahydrofuran & 5 \\
1-Aminopentane & 5 \\
Acrylonitrile & 5 \\
2-Butanone & 5 \\
1,4-Dioxane & 5 \\
Acetone & 5 \\
Methyl formate & 5 \\
Cyclopentanone & 5 \\
Formic acid ethyl ester & 5 \\
Methyl acetate & 5 \\
1,3-Cyclopentadiene & 5 \\
2-Pentanone & 5 \\
Methyl isopropyl ketone & 5 \\
Propyl acetate & 5 \\
3-Pentanone & 5 \\
Propanoic acid ethyl ester & 5 \\
Acetic acid isopropyl ester & 5 \\
Formic acid propyl ester & 5 \\
Ethyl acetate & 5 \\
Diisobutyl ether & 5 \\
Dipentyl ether & 5 \\
Diethyl ether & 5 \\
Hexylamine & 5 \\
Dipropylamine & 5 \\
Carbon dioxide & 5 \\
Methyl iodide & 5 \\
1,3-Cyclohexadiene & 5 \\
Chloroform & 5 \\
Dichloromethane & 5 \\
Pyridine & 5 \\
Butyraldehyde & 6 \\
Octanal & 6 \\
sec-Butylamine & 6 \\
N,N-Diethylamine & 6 \\
Tributylamine & 6 \\
Acetaldehyde & 6 \\
Butylamine & 6 \\
Propanal & 6 \\
Ethanol & 6 \\
Methanol & 6 \\
Acetic acid & 6 \\
N,N-Dimethylformamide (DMF) & 6 \\
N-Methyl-2-pyrrolidone & 6 \\
Dimethyl sulfoxide & 6 \\
Cyclohexanone & 6 \\
N,N-Dimethylacetamide & 6 \\
N,N-Dimethyl propanoic acid amide & 6 \\
Propane & 7 \\
n-Butane & 7 \\
2-Methylpropane & 7 \\
Hexyl acetate & 7 \\
Cyclohexene & 7 \\
1-Pentene & 7 \\
2-Methyl-2-butene & 7 \\
Cyclopentane & 7 \\
Dibutyl ether & 7 \\
3-Methyl-1-butene & 7 \\
Nitrobenzene & 7 \\
2-Heptanone & 7 \\
Isoamyl acetate & 7 \\
1-Octen-3-ol & 7 \\
Diisobutyl ketone & 7 \\
Climbazole & 7 \\
1,1,2,2-Tetrachloroethane & 7 \\
Hexanal & 7 \\
Triethylamine & 8 \\
2-Methyl-2-pentene & 8 \\
trans-2-Pentene & 8 \\
Isobutylene & 8 \\
Ethyl tert-butyl ether (ETBE) & 8 \\
Tetrahydropyran & 8 \\
Methyl tert-butyl ether (MTBE) & 8 \\
1,3-Butadiene & 8 \\
Valeraldehyde & 8 \\
3-Methylpyridine & 8 \\
2-Methyl-1-butene & 8 \\
Triacontane & 8 \\
Diisopropyl ether & 8 \\
2-Hexanone & 8 \\
Ethyl butyrate & 8 \\
3-Heptanone & 8 \\
Butyl acetate & 8 \\
Isobutyl acetate & 8 \\
Dibromomethane [R30B2] & 8 \\
Trichloroethylene & 8 \\
Cyclopentene & 8 \\
1,4-Cyclohexadiene & 8 \\
1-Octadecyl naphthalene & 8 \\
1,1-Dichloroethane [R150a] & 8 \\
tert-Butyl chloride & 8 \\
Carbon disulfide & 8 \\
1,1,2-Trichloroethane & 8 \\
Di-n-propyl ether & 8 \\
Tripropylamine & 8 \\
1-Octanamine & 8 \\
1-Dodecyl decahydronaphthalene & 8 \\
1-Butene & 8 \\
Methyl tert-amyl ether (TAME) & 8 \\
Heptylamine & 8 \\
Anthracene & 9 \\
Phenanthrene & 9 \\
1,3,5-Trimethylbenzene & 9 \\
1-Methylnaphthalene & 9 \\
Tetrachloroethylene & 9 \\
1,5-Hexadiene & 9 \\
4-Ethenylcyclohexene & 9 \\
1-Chloropentane & 9 \\
Butylbenzene & 9 \\
tert-Butylbenzene & 9 \\
1,7-Octadiene & 9 \\
4-Isopropyltoluene & 9 \\
Limonene & 9 \\
Naphthalene & 10 \\
Biphenyl & 10 \\
1,2-Epoxy-p-menth-8-ene & 10 \\
Ethyl iodide & 10 \\
Toluene & 10 \\
Isopropyl bromide & 10 \\
Amyl acetate & 10 \\
Butyl chloride & 10 \\
Ethylbenzene & 10 \\
1,1,1-Trichloroethane [R140a] & 10 \\
o-Xylene & 10 \\
Tetrachloromethane & 10 \\
Propyl bromide & 10 \\
m-Xylene & 10 \\
p-Xylene & 10 \\
Isoprene & 10 \\
trans-1,3-Pentadiene & 10 \\
cis-1,3-Pentadiene & 10 \\
1,2,3,4-Tetrahydronaphthalene & 10 \\
1-Hexyne & 10 \\
1-Heptyne & 10 \\
Cycloheptatriene & 10 \\
1,3-Butadiene, 2,3-dimethyl- & 10 \\
Isopropylbenzene & 10 \\
Propylbenzene & 10 \\
1-Octyne & 10 \\
Benzyl chloride & 10 \\
Benzyl bromide & 10 \\
Thiophene & 11 \\
Ethyl bromide & 11 \\
Benzene & 11 \\
1-Chloropropane & 11 \\
Dimethyl sulfide & 11 \\
4-Methyl-2-pentanone & 11 \\
Furan & 11 \\
1,2-Dichloropropane & 11 \\
Fluorobenzene & 11 \\
Chlorobenzene & 11 \\
1,2-Dichloroethane & 11 \\
Benzaldehyde & 11 \\
Methoxybenzene & 11 \\
1-Pentyne & 11 \\
Bromobenzene & 11 \\
o-Dichlorobenzene & 11 \\

\end{longtable}

\begin{longtable}{ll}
\caption{DDB name and class number of solvents, obtained from hierarchical clustering on a matrix predicted by sMCM.} \label{tab:long} \\
\hline
\textbf{DDB Name} & \textbf{Class Number} \\
\hline
\endfirsthead

\multicolumn{2}{c}%
{{\bfseries \tablename\ \thetable{} -- continued from previous page}} \\
\hline
\textbf{DDB Name} & \textbf{Class Number} \\
\hline
\endhead

\hline \multicolumn{2}{r}{{Continued on next page}} \\ \hline
\endfoot

\hline \hline
\endlastfoot

Water & 0 \\
Deuterium oxide <Heavy water> & 0 \\
1,2-Ethanediol & 1 \\
Formamide & 1 \\
1,1,1,3,3,3-Hexafluoro-2-propanol & 2 \\
Perfluorotributylamine & 2 \\
Perfluoro-n-octane & 2 \\
1-Butanol & 2 \\
1-Propanol & 2 \\
Ethanol & 2 \\
2-Propanol & 2 \\
N-Methyl-2-piperidone & 2 \\
2,6-Dimethoxyphenol & 2 \\
3-Methyl sulfolane & 2 \\
1,5-Dicyanopentane & 2 \\
N,N-Dimethylacetamide & 2 \\
N-Methyl-2-pyrrolidone & 2 \\
Furfural & 2 \\
N,N-Dimethylformamide (DMF) & 2 \\
Trimethyl phosphate & 2 \\
Aniline & 2 \\
Acetic acid & 2 \\
N-Isopropylacetamide & 2 \\
2-Ethoxyethanol & 2 \\
alpha-Aminotoluene & 2 \\
N,N-Diethylacetamide & 2 \\
2,5-Hexanedione & 2 \\
Benzylcyanide & 2 \\
2-Methoxyethanol & 2 \\
N-Methylacetamide & 2 \\
Diethylene glycol monomethyl ether & 2 \\
Tetrahydrofurfuryl alcohol & 2 \\
Dichloroacetic acid & 2 \\
Dimethylcyanamide & 2 \\
2,4-Dimethylsulfolane & 2 \\
Phenol & 2 \\
1,3-Dimethylimidazolidin-2-one & 2 \\
Benzyl alcohol & 2 \\
N-Isopropylformamide & 2 \\
Triethylene glycol & 3 \\
3,3'-Oxybispropionitrile & 3 \\
Dimethyl sulfoxide & 3 \\
Propylene carbonate & 3 \\
2-Pyrrolidone & 3 \\
Sulfolane & 3 \\
N-Formylmorpholine & 3 \\
Tetraethylene glycol & 3 \\
Methyleneglutaronitrile & 3 \\
Divinylsulfone & 3 \\
Methylglutaronitrile & 3 \\
1,4-Dicyanobutane & 3 \\
gamma-Butyrolactone & 3 \\
Ethylene sulfite <Glycol sulfite> & 3 \\
Furfuryl alcohol & 3 \\
Ethylene carbonate & 3 \\
Acetonitrile & 3 \\
Nitromethane & 3 \\
4-Chloromethyl-2-one-1,3-dioxolane & 3 \\
Glutaronitrile & 3 \\
Tetramethylene sulfoxide & 3 \\
Iminodipropionitrile & 3 \\
1,4-Butanediol & 3 \\
2-Mercapto ethanol & 3 \\
2-Chloroethanol & 3 \\
1,5-Pentanediol & 3 \\
N-Methylformamide & 3 \\
beta-Chloropropionitrile & 3 \\
Diethylene glycol & 3 \\
Ethylenediamine & 3 \\
Ethylene cyanohydrin & 3 \\
1,3-Propanediol & 3 \\
1,2-Propanediol & 3 \\
2,2'-Diethyl-dihydroxy sulfide & 3 \\
Methanol & 3 \\
1,6-Hexanediol & 3 \\
2,2,2-Trifluoroethanol & 3 \\
Mono-n-butyl phosphate & 3 \\
Monoethanolamine & 3 \\
N-Methylmethansulfonamide & 3 \\
1,2-Dicyanoethane & 3 \\
Malonic acid dinitrile & 3 \\
Glycerol & 3 \\
Fumaronitrile & 3 \\
Maleonitrile & 3 \\
Bis(2-ethylhexyl) phosphate & 4 \\
Indene & 5 \\
3-Pentanone & 5 \\
Carbonic acid diethyl ester & 5 \\
2-Methylpyridine & 5 \\
1-Methylnaphthalene & 5 \\
Ethyl benzoate & 5 \\
Cyclohexanone & 5 \\
Phthalic acid benzyl butyl ester & 5 \\
Diethylene glycol diethyl ether & 5 \\
Cyclohexyl acetone & 5 \\
1,1,2,2-Tetrachloroethane & 5 \\
Octanenitrile & 5 \\
Heptanenitrile & 5 \\
Bis(2-ethylhexyl) phthalate & 5 \\
Amyl acetate & 5 \\
Tetrahydrofuran & 5 \\
Phthalic acid dibutyl ester & 5 \\
Propyl acetate & 5 \\
Nonanenitrile & 5 \\
1,2,3,4-Tetrahydronaphthalene & 5 \\
Pentadecanoic acid, nitrile & 5 \\
Phthalic acid dinonyl ester & 5 \\
Bis-(2-ethylhexyl)-sebacate & 5 \\
Di(2-ethylhexyl) adipate & 5 \\
Tripentylamine & 5 \\
Trioctylamine & 5 \\
Trihexylamine & 5 \\
N,N-Dibutyl-2-ethylhexylamide & 5 \\
N,N-Dibutyl-2,2-dimethylbutanamide & 5 \\
N,N-Diethyl dodecanamide & 5 \\
Tributyl phosphate & 5 \\
Tris-butoxyethyl phosphate & 5 \\
Tributylamine & 5 \\
Triethyl phosphate & 5 \\
Linoleic acid & 5 \\
2,6-Dimethylpyridine & 5 \\
1,1,1-Trichloroethane [R140a] & 5 \\
Chlorocyclohexane & 5 \\
Bromocyclohexane & 5 \\
cis-1,2-Dichloroethylene & 5 \\
trans-1,2-Dichloroethene & 5 \\
1-Octene & 5 \\
1-Decene & 5 \\
Ethyl tert-butyl ether (ETBE) & 5 \\
Butylphenol & 6 \\
Amylphenol & 6 \\
Hexamethylphosphoric acid triamide & 6 \\
1,3-Dimethoxybenzene <Resorcinol dimethyl ether> & 6 \\
2,4-Diisopropylphenol & 6 \\
Perfluorohexane & 7 \\
Perfluoro-n-heptane & 7 \\
Diiodomethane & 7 \\
Carbon disulfide & 8 \\
Cyclohexane & 8 \\
Pentane & 8 \\
Decane & 8 \\
Hexane & 8 \\
2,2,4-Trimethylpentane & 8 \\
Heptane & 8 \\
Octane & 8 \\
Dodecane & 8 \\
Nonane & 8 \\
Dibutyl ether & 8 \\
Tetradecane & 8 \\
Squalane & 8 \\
Tetrachloromethane & 8 \\
Hexadecane & 8 \\
Diisopropyl ether & 9 \\
Diethyl ether & 9 \\
1-Hexene & 9 \\
Triethylamine & 9 \\
Phenylcyclohexane & 9 \\
Acetanilide & 9 \\
Hexyl acetate & 9 \\
Octamethylcyclotetrasiloxane & 9 \\
Bicyclohexyl & 9 \\
trans-Decahydronaphthalene & 9 \\
cis-Decahydronaphthalene & 9 \\
Hexafluorobenzene & 10 \\
Iodobenzene & 10 \\
Ethyl bromide & 10 \\
Chloroform & 10 \\
Ethoxybenzene & 10 \\
Butoxybenzene & 10 \\
p-Xylene & 10 \\
Benzene & 10 \\
Toluene & 10 \\
Fluorobenzene & 10 \\
Bromobenzene & 10 \\
Chlorobenzene & 10 \\
Acetophenone & 11 \\
Benzonitrile & 11 \\
Phthalic acid diethyl ester & 11 \\
Ethyl acetate & 11 \\
Propyl phenyl ketone & 11 \\
Benzophenone & 11 \\
Salicylic acid methyl ester & 11 \\
4-Bromoanisole & 11 \\
1-Bromonaphthalene & 11 \\
2-Butoxyethanol & 11 \\
1,2-Epoxy-p-menth-8-ene & 11 \\
Dichloromethane & 11 \\
1,1-Dichloroethane [R150a] & 11 \\
Methoxybenzene & 11 \\
Dibenzyl ether & 11 \\
Limonene & 12 \\
1,1,2,2-Tetrabromoethane & 12 \\
Nitroethane & 12 \\
Propionitrile & 12 \\
Nitrobenzene & 12 \\
1-(1-Naphthalenyl)ethanone & 12 \\
Methyl acetate & 13 \\
Cycloheptanol & 13 \\
Cyclopentanol & 13 \\
Cyclohexanol & 13 \\
tert-Butanol & 13 \\
3-Methyl-1-butanol & 13 \\
1-Pentanol & 13 \\
1-Hexanol & 13 \\
Tetraethylene glycol dimethyl ether & 13 \\
Di-n-butyl phosphate & 14 \\
3-Methylphenol & 14 \\
N-Ethylpropionamide & 14 \\
N-Acetyloxazolidine & 14 \\
N-Methylisobutyramide & 14 \\
Diethyl oxalate & 14 \\
N-Ethylacetamide & 14 \\
N-Methyl propanamide & 14 \\
2-Phenylethanol & 14 \\
Acetone & 14 \\
Butanenitrile & 14 \\
2-Isopropoxyethanol & 14 \\
Ethylene glycol monopropyl ether & 14 \\
2,4-Pentanedione & 14 \\
N-Acetylpiperidine & 14 \\
1-Octanol & 15 \\
1-Dodecanol & 15 \\
Butyl acetate & 15 \\
2-Heptanone & 15 \\
Cyclopentanone & 15 \\
2-Pentanone & 15 \\
1-Heptanol & 15 \\
tert-Pentanol & 15 \\
Ricinoleic acid & 15 \\
Ethyl butyrate & 15 \\
1,2-Dichloroethane & 16 \\
Acetic acid benzyl ester & 16 \\
1-Chloronaphthalene & 16 \\
Hexanenitrile & 16 \\
1,4-Dioxane & 16 \\
Quinoline & 16 \\
2-Butanone & 16 \\
Ethyl phenyl ketone & 16 \\
Phenylacetone & 16 \\
1,6-Dicyanohexane & 16 \\
N,N-Dimethyl propanoic acid amide & 16 \\
N,N-Dimethylisobutyramide & 16 \\
1,5-Dimethyl-2-pyrrolidone & 16 \\
N-Ethyl-2-pyrrolidone & 16 \\
Pyridine & 16 \\
1-Nitropropane & 16 \\
N-Methylcaprolactam & 16 \\
4-Phenyl-2-butanone & 16 \\
Methyl diphenyl phosphate & 16 \\
Pentanenitrile & 16 \\
\end{longtable}